\documentclass[letterpaper, preprint, paper,11pt]{AAS}	% for preprint proceedings

\usepackage{bm}
\usepackage{amsmath}
\usepackage{subfigure}
\usepackage[colorlinks=true, pdfstartview=FitV, linkcolor=black, citecolor= black, urlcolor= black]{hyperref}
\usepackage{overcite}
\usepackage{footnpag}	
\usepackage{xfrac}
\usepackage{mathdots}
\usepackage{relsize}
\usepackage{subcaption}
\usepackage{mathtools, nccmath}
\usepackage{amsfonts}
\usepackage{acronym}

\usepackage{array}
\usepackage{comment}
\usepackage{dsfont}
\usepackage{amssymb}

\usepackage{amsthm}

\newacro{KLD}{Kullback-Leibler Divergence}
\newacro{SCvx}{Successive Convexification}
\newacro{SCP}{Sequential Convex Programming}
\newacro{DRO}{Distant Retrograde Orbit}
\newacro{NRHO}{Near-Rectilinear Halo Orbit}
\newacro{CRTBP}{Circular Restricted Three Body Problem}
\newacro{CRLB}{Cramer-Rao Lower Bound}
\newacro{RMS}{Root Mean Square}
\newacro{SSA}{Space Situational Awareness}
\newacro{SDA}{Space Domain Awareness}
\newacro{DSN}{Deep Space Network}
\newacro{SSAT}{Spacecraft-to-Spacecraft Absolute Tracking}
\newacro{PSD}{power spectral density}
\newacro{GMM}{Gaussian Mixture Model}
\newacro{SDE}{Stochastic Differential Equation}
\newacro{MSE}{Mean Squared Error}
\newacro{CVaR}{Conditional Value at Risk}
\newacro{VaR}{Value at Risk}
\newacro{CDF}{Cummulative Distribution Function}
\newacro{FOH}{First-Order Hold}
\newacro{PDF}{Probability Distribution Function}

\PaperNumber{26-978}

\begin{document}

\title{Approximate Relative Entropy Constraints for Nonlinear Covariance Steering Under Distribution Ambiguity}

\author{Trevor N. Wolf\thanks{Postdoctoral Associate, Department of Aerospace Engineering Sciences, University of Colorado Boulder, AAS Member.},  
Jay W. McMahon\thanks{Associate Professor, University of Colorado Boulder, Department of Aerospace Engineering Sciences, and AAS Fellow.}
}

\maketitle{} 		

\begin{abstract}
Covariance steering provides an efficient framework for designing linear stochastic feedback policies, but its extension to nonlinear systems relies on a Gaussian surrogate obtained through local linearization. Because this surrogate may differ substantially from the true nonlinear state distribution, risk-sensitive quantities such as collision probability and mean-squared error may be inaccurately estimated. This work develops a distributionally robust covariance-steering framework based on the relative entropy, also known as the Kullback–Leibler divergence (KLD), to account for ambiguity in the propagated probability density function. Using a variational representation of exponential integrals, we derive computable upper bounds on risk-sensitive quantities over a KLD ambiguity set. We then formulate an upper bound on the time rate of change of the KLD between the true nonlinear distribution and a Gaussian reference surrogate. Under some assumptions, this bound is controlled by decision variables within a covariance-steering formulation. The resulting constraints are incorporated into a sequential convex programming algorithm to design stochastic guidance policies that keep the true distribution close to its Gaussian surrogate while enforcing bounds on risk-sensitive performance measures. The proposed approach is demonstrated on a challenging nonlinear spacecraft transfer between two near-rectilinear halo orbits.
\end{abstract}

% ----------------------------------
% ---------- Introduction ----------
% ----------------------------------

\section{Introduction}

Spacecraft dynamics are subject to uncertainty and random errors that must be accounted for in designing guidance feedback policies. Rather than steering a deterministic trajectory, stochastic guidance seeks to steer a probability distribution of the dynamical state to, or within, a desired terminal \ac{PDF}. Gaussian covariance steering has become an attractive tool for forming linear-affine stochastic control policies, both within the aerospace and among many other engineering disciplines \cite{Okamoto_2018, Oguri_2021, Chen_2021, Liu_2024b}. For linear Gaussian systems, the covariance steering problem can be formulated as a semidefinite program that is solvable with many open-source and commercial convex programming packages \cite{Liu_2024}. The technique can be extended to nonlinear systems by linearizing the dynamics around a nominal reference trajectory and assuming initial Gaussian uncertainty. This extension has seen considerable attention in the spaceflight community. However, because the resulting Gaussian \ac{PDF} is only an approximation of the true underlying stochastic state of the nonlinear system, a user must be careful in placing trust in this result. This is particularly concerning in evaluating and constraining expected risk-sensitive quantities using the resulting surrogate Gaussian distribution. 

Recent work has explored using higher-fidelity \ac{PDF} approximations with stochastic spacecraft guidance that should more accurately represent the true underlying state uncertainty. These include using the conjugate unscented transform \cite{Qi_2025}, or \acp{GMM} \cite{Boone_2022, Kumagai_2024} for uncertainty propagation. Alternatively, a user may be interested in including uncertainty in the \ac{PDF} itself when designing guidance feedback, a technique known as distributionally robust control \cite{Erdogan_2006}. With the latter, a guidance policy is formulated that is robust to \acp{PDF} residing within an ambiguity set of plausible \acp{PDF} for the stochastic spacecraft state. This set is found by bounding a statistical distance measure with respect to a reference \ac{PDF}. The Wasserstein distance is one of these measures, and several recent works have explored its use in designing robust guidance policies for linear time-invariant systems under distributional ambiguity \cite{Aolaritei_2023, Pilipovsky_2024}. Regardless of the metric used to construct the ambiguity set, a key unresolved challenge is the principled selection of an appropriate ambiguity-set radius.

In this work, we use the relative entropy, also known as the \ac{KLD}, to form an ambiguity set for a spacecraft's state \ac{PDF}, and provide a principled approach for determining an appropriate bound radius. As an aside, the reader should note that the terms relative entropy and \ac{KLD} shall be used interchangeably throughout. The choice of the \ac{KLD} as a distance measure in this context is advantageous for its numerous connections to dynamical systems theory and large deviation theory. To that end, we show, through a variational expression for exponential integrals, that key quantities of interest can be upper-bounded, for instance, the collision risk and the \ac{MSE}. Next, we formulate an approach for bounding the time rate of change of the relative entropy whose integrated state is compatible with computing the aforementioned risk-sensitive variational expression. Under some assumptions, namely that the reference surrogate is Gaussian,  this \ac{KLD} upper bound is controllable within a covariance steering framework. Finally, using the \ac{SCvx} algorithm, we incorporate these analytical results to design robust stochastic guidance policies that can 1) constrain the resulting true uncertainty to remain close to its reference surrogate over time, and 2) constrain the upper bound on risk-sensitive quantities, and thus, those quantities themselves. Our methods are tested on a challenging nonlinear guidance problem involving a transfer trajectory between two \acp{NRHO}.

% ------------------------------------------
% ---------- Problem Formulation -----------
% ------------------------------------------

\section{Preliminaries}

This section describes common model and structural approximations used in stochastic spacecraft guidance. These approximations are relevant to the well-studied covariance steering approach that we base our methods on later. The true stochastic dynamics of the spacecraft's state are prescribed by the \ac{SDE} 
\begin{equation}
    d\boldsymbol{x}_t = \boldsymbol{\phi}_t(\boldsymbol{x}_t, \boldsymbol{u}_t)dt + G_td\boldsymbol{w}_t, \hspace{1ex} \boldsymbol{x}_0 \sim \pi_0.
\end{equation}
We use the subscript $t$ to denote time dependence. The function $\boldsymbol{\phi}_t(\cdot)$ represents the \textit{true} nonlinear drift dynamics of the system. The distribution $\pi_0$ captures solely aleatoric uncertainty in the initial state distribution. That is to say, $\pi_0$ is the \textit{true} initial state distribution of the spacecraft. The matrix $G_t$ is the process noise gain matrix, and $d\boldsymbol{w}_t \sim \mathcal{N}(0_{n\times1}, Qdt)$ where $Q$ is the process noise \ac{PSD}. 

A user does not have complete knowledge of $\boldsymbol{\phi}_t(\cdot)$ and $\pi_0$, but rather substitutes a model for the drift, which we call $\boldsymbol{f}_t(\cdot)$, and, often, an initial Gaussian state distribution. Furthermore, for tractability, structural approximations, often in the form of linearization, are used. With this, to design guidance feedback, we assume a reference \ac{SDE} as
\begin{equation}
    d\boldsymbol{x}_t = \hat{\boldsymbol{f}}_t(\boldsymbol{x}_t, \boldsymbol{u}_t) + G_td\boldsymbol{w}_t, \hspace{1ex} \boldsymbol{x}_0 \sim \mathcal{N}(\boldsymbol{x}_0; \boldsymbol{\mu}_0, P_0),  
\end{equation}
The approximate drift is an affine function, 
\begin{equation}
    \hat{\boldsymbol{f}}_t(\boldsymbol{x}_t, \boldsymbol{u}_t) = A_t\boldsymbol{x}_t + B_t\boldsymbol{u}_t + \boldsymbol{r}_t, 
\end{equation}
where, 
\begin{equation}
    A_t = \nabla_{\boldsymbol{x}_t} \boldsymbol{f}_t(\boldsymbol{\mu}_t, \boldsymbol{u}_t),   
\end{equation}
\begin{equation}
    \boldsymbol{r}_t = \boldsymbol{f}_t(\boldsymbol{\mu}_t, \boldsymbol{u}_t) - A_t\boldsymbol{\mu}_t - B_t \boldsymbol{u}_t, 
\end{equation}
and the matrix $B_t$ is the control gain matrix. 

Under these linearized dynamics, provided the initial distribution is Gaussian, it will remain so. This feature, however, is inconsistent with our nonlinear model $\boldsymbol{f}_t(\cdot)$ whose action on $\pi_0$ is such that the distribution becomes non-Gaussian over time. The non-Gaussian evolution, and imprecise model parameters, introduces what we classify as distribution ambiguity into stochastic guidance design. To be explicit, by steering the designed-for Gaussian distribution using linearized stochastic guidance techniques, we can not ensure consistentcy because the policy design lacks knowledge of the true underlying distribution. This is especially problematic if a user desires to place guarantees on expected quantities of interest -- for instance, probability of collision or \ac{MSE}. In the following sections, we introduce approaches to handle this type of ambiguity in stochastic guidance design. 

% ----------------------------------
% ---------- Methodology -----------
% ----------------------------------

\section{Methodology}

% -------------------------------------------------- %
% ------------ Constraining Quantities ------------- %
% -------------------------------------------------- %

\subsection{Constraining Expected Quantities Under Distribution Ambiguity}

We use a variational expression for exponential integrals, otherwise known as the Donsker-Varadhan formula \cite{Dupuis_1997, Dembo_1998}, to bound expected quantities of interest under distribution ambiguity. Let $\pi$ be an exact but unknown or intractable probability distribution for the random variable $\boldsymbol{x}$, and $\hat{\pi}$ be a surrogate approximation of $\pi$ available to the user. For any bounded function $g(\cdot)$,
\begin{equation}
    \frac{1}{\lambda}
    \ln\!\left(
    \mathbb{E}_{\hat{\pi}}\!\left[e^{\lambda g(\boldsymbol{x})}\right]
    \right)
    =
    \sup_{\pi \in \mathcal{P}(\mathcal{X})}
    \left[
    -\frac{1}{\lambda}\mathcal{R}(\pi\|\hat{\pi})
    +
    \mathbb{E}_{\pi}\!\left[g(\boldsymbol{x})\right]
    \right].\label{eqn:DV_formula}
\end{equation}
The variable $\lambda$ is a positive scalar and $\mathcal{R}(\pi\|\hat{\pi})$ denotes the relative entropy, also known as the \ac{KLD} \cite{Kullback_1951}, defined as
\begin{equation}
    \mathcal{R}(\pi\|\hat{\pi}) = \int_{\mathcal{X}}\pi(\boldsymbol{x}) \ln \left( \frac{\pi(\boldsymbol{x})}{\hat{\pi}(\boldsymbol{x})} \right) d\boldsymbol{x}. \label{eqn:kld}
\end{equation}
From Eq. \ref{eqn:DV_formula}, it follows that
\begin{equation}
    \mathbb{E}_{\pi}[g(\boldsymbol{x})] \leq \frac{1}{\lambda}\left\{\ln\left(\mathbb{E}_{\hat{\pi}}\left[e^{\lambda g(\boldsymbol{x})}\right] \right) + \mathcal{R}(\pi\|\hat{\pi}) \right\}. \label{eqn:general_upper_bound}
\end{equation}
Notice that bounding the left-hand side requires only an expectation evaluated under the known surrogate distribution, $\hat{\pi}$, and an evaluation of the statistical distance, $\mathcal{R}(\pi_t\|\hat{\pi}_t)$, between $\pi$ and $\hat{\pi}$. 

The expression above presents several useful applications. For instance, in stochastic spacecraft guidance, it is often desirable to constrain expected quantities of interest including risk probability, \ac{VaR}, \ac{CVaR}, and delivery \ac{MSE} \cite{Ridderhof_2020, Ridderhof_2022}. However, in the absence of an exact distribution $\pi$, constraining these quantities loses rigorous guarantees. Using Eq. \ref{eqn:general_upper_bound}, the following section derives explicit bounds for two of these quantities, namely the collision risk and \ac{MSE}, that do not require knowledge of an exact distribution under which these expected quantities should be evaluated. 

% --------------------------------------------- %
% ------------ Chance Constraints ------------- %
% --------------------------------------------- %

\subsubsection{Chance Constraint Upper Bound:}

Let the time-evolving set $\mathcal{S}_t$ represent the admissible dynamical states of a spacecraft. This can, for instance, exclude regions that would risk collision with a central body or other space objects. A chance constraint evaluated under the true distribution $\pi_t$ is then 
\begin{equation}
\mathbb{P}_{\boldsymbol{x}_t \sim \pi_t}
\!\left[\boldsymbol{x}_t \in \mathcal{S}_t\right]
\geq 1-\epsilon
\quad \Longleftrightarrow \quad
\mathbb{P}_{\boldsymbol{x}_t \sim \pi_t}
\!\left[\boldsymbol{x}_t \notin \mathcal{S}_t\right]
< \epsilon,
\end{equation}
where $\epsilon$ is a small admissible risk probability (e.g., 0.01). We can write the risk probability as an expectation of $g(\boldsymbol{x}_t) = 1 - \mathds{1}_{\{\boldsymbol{x}_t \in \mathcal{S}_t \}}$,
\begin{equation}
\mathbb{P}_{\boldsymbol{x}_t \sim \pi_t}
\!\left[
\boldsymbol{x}_t \notin S_t
\right]
=
\mathbb{E}_{\pi_t}
\!\left[
1-\mathds{1}_{\{\boldsymbol{x}_t \in S_t\}}
\right],
\end{equation}
where $\mathds{1}_{\{ \cdot \}}$ is the indicator function. Then, using Eq. \ref{eqn:general_upper_bound}, 
\begin{flalign}
    \mathbb{P}_{\boldsymbol{x}_t \sim \pi_t}\!\left[\boldsymbol{x}_t \notin S_t\right]
&\leq
\frac{1}{\lambda_t}
\left\{
\ln\!\left(
\mathbb{E}_{\hat{\pi}_t}\!\left[
e^{\lambda_t\left(1-\mathds{1}_{\{\boldsymbol{x}_t \in S_t\}}\right)}
\right]
\right)
+
\mathcal{R}\!\left(\pi_t\middle\|\hat{\pi}_t\right)
\right\} \nonumber\\
&=
1+\frac{1}{\lambda_t}
\left\{
\ln\!\left(
1-\left(1-e^{-\lambda_t}\right)
\mathbb{P}_{\boldsymbol{x}_t \sim \hat{\pi}_t}\!\left[\boldsymbol{x}_t \in S_t\right]
\right)
+
\mathcal{R}\!\left(\pi_t\middle\|\hat{\pi}_t\right)
\right\}.
\end{flalign}
Assume the surrogate distribution is Gaussian, $\hat{\pi}_t \sim \mathcal{N}(\boldsymbol{x}_t; \boldsymbol{x}_t^*, P_t)$, where, in this work \footnote{The surrogate distribution does not need to be Gaussian in general.}, $\boldsymbol{x}^*_t$ and $P_t$ are the target mean and covariance generated with covariance steering. Furthermore, if the admissible set $\mathcal{S}_t$ is a half-space, i.e., 
\begin{equation}
    \mathcal{S}_t=\left\{\boldsymbol{x}_t\in\mathbb{R}^n\mid \boldsymbol{a}_t^{\top}\boldsymbol{x}_t\leq b_t\right\}, 
\end{equation}
then
\begin{equation}
    \mathbb{P}_{\boldsymbol{x}_t \sim \pi_t}\!\left[\boldsymbol{x}_t \notin \mathcal{S}_t\right]
    \leq
    1+\frac{1}{\lambda_t}
    \left\{
    \ln\!\left(
    1-\left(1-e^{-\lambda_t}\right)
    \operatorname{cdf}\!\left(
    \frac{b_t-\boldsymbol{a}_t^{\top}\boldsymbol{x}^*_t}
    {\sqrt{\boldsymbol{a}_t^{\top}P_t\boldsymbol{a}_t}}
    \right)
    \right)
    +
    \mathcal{R}\!\left(\pi_t\middle\|\hat{\pi}_t\right)
    \right\}. \label{eqn:upper_bound_risk}
\end{equation}
The function $\operatorname{cdf}(\cdot)$ is the standard normal \ac{CDF}. There is a unique minimizer $\lambda_t^*$ for the expression given by Eq. \ref{eqn:upper_bound_risk}. It can be shown that \footnote{ See Ref. \citenum{Wolf_2026}.} as $\mathcal{R}(\pi_t\| \hat{\pi}_t) \rightarrow 0$, the minimizer $\lambda^* \rightarrow 0$, and 
\begin{flalign}
    \lim_{\mathcal{R}(\cdot), \lambda_t^{*}\to 0}
    &1+\frac{1}{\lambda_t^{*}}
    \left\{
    \ln\!\left(
    1-\left(1-e^{-\lambda_t^{*}}\right)
    \operatorname{cdf}\!\left(
    \frac{b_t-\boldsymbol{a}_t^{\top}\boldsymbol{x}^*_t}
    {\sqrt{\boldsymbol{a}_t^{\top}P_t\boldsymbol{a}_t}}
    \right)
    \right)
    +
    \mathcal{R}\!\left(\pi_t\middle\|\hat{\pi}_t\right)
    \right\}\nonumber\\
    &=
    1-\operatorname{cdf}\!\left(
    \frac{b_t-\boldsymbol{a}_t^{\top}\boldsymbol{x}^*_t}
    {\sqrt{\boldsymbol{a}_t^{\top}P_t\boldsymbol{a}_t}}
    \right).\label{eqn:gaussian_risk}
\end{flalign}
In other words, as the relative entropy converges to zero, the true distribution and its Gaussian approximation become identical. Therefore, the risk probability is the exact value evaluated under the Gaussian surrogate and provided by the second expression in Eq. \ref{eqn:gaussian_risk}.

% ---------------------------------------------- %
% ------------- Mean Squared Error ------------- %
% ---------------------------------------------- %

\subsubsection{\ac{MSE} Upper Bound:} Let $g(\boldsymbol{x}_t) = \tilde{\boldsymbol{x}}_t^\top \tilde{\boldsymbol{x}}_t$, where $\tilde{\boldsymbol{x}} = \boldsymbol{x}_t - \boldsymbol{x}_t^*$. Then the true delivery \ac{MSE} at time $t$ is 
\begin{equation}
    \text{MSE}_t = \mathbb{E}_{\pi_t}[\tilde{\boldsymbol{x}}^\top \tilde{\boldsymbol{x}}_t].
\end{equation}
An upper bound for this quantity follows from Eq. \ref{eqn:general_upper_bound}, written as 
\begin{equation}
    \mathbb{E}_{\pi_t}[\tilde{\boldsymbol{x}}^\top\tilde{\boldsymbol{x}}] \leq \frac{1}{\lambda_t}\left\{\ln\left(\mathbb{E}_{\hat{\pi}_t}\left[e^{\lambda_t\tilde{\boldsymbol{x}}^\top \boldsymbol{\tilde{x}}} \right] \right) + \mathcal{R}(\pi_t\|\hat{\pi}_t) \right\}.\label{eqn:MSE_general_ub}
\end{equation}
Using our convention that the surrogate distribution, $\hat{\pi}_t$, is Gaussian distributed with mean $\boldsymbol{x}_t^*$ and covariance $P_t$, the expectation in Eq. \ref{eqn:MSE_general_ub} is   
\begin{equation}
    \mathbb{E}_{\hat{\pi}_t}[e^{\lambda_t\tilde{\boldsymbol{x}}_t^\top\tilde{\boldsymbol{x}}_t}] = \operatorname{det}(I - 2\lambda_t P_t)^{-1/2}, \hspace{1ex} \text{s.t. } I - 2\lambda_t P_t \succ 0. 
\end{equation}
It is fairly easy to show then
\begin{equation}
    \mathbb{E}_{\pi_t}[\tilde{\boldsymbol{x}}_t^\top \tilde{\boldsymbol{x}}_t] \leq \frac{1}{\lambda_t}\left\{-\frac{1}{2} \operatorname{logdet}(I - 2\lambda_t P_t) + \mathcal{R}(\pi_t\| \hat{\pi}_t) \right\}.\label{eqn:MSE_upper_bound}
\end{equation}
The function $\operatorname{logdet}(\cdot)$ is the natural logarithm of the determinant of a matrix. As the distance $\mathcal{R}(\pi_t\|\hat{\pi}_t) \rightarrow 0$, the minimizer $\lambda^*_t \rightarrow 0$, and it can be shown that
\begin{flalign}
\lim_{\mathcal{R}(\cdot), \lambda_t^* \rightarrow 0} &\frac{1}{\lambda_t}\left\{-\frac{1}{2} \operatorname{logdet}(I - 2\lambda_t P_t) + \mathcal{R}(\pi_t\| \hat{\pi}_t) \right\} \nonumber \\
& = \operatorname{tr}(P_t),
\end{flalign}
which is the exact \ac{MSE} evaluated under the Gaussian surrogate.

We have shown that expected quantities evaluated under an intractable distribution $\pi_t$ can be bounded, provided a user has knowledge of the statistical distance between it and the surrogate distribution, $\hat{\pi}_t$. A natural question then is how a user can determine this distance. Without knowledge of $\pi_t$, it is not be possible to determine $\mathcal{R}(\pi_t\| \hat{\pi}_t)$ exactly. However, under some assumptions, the following section derives a useful upper bound for this distance that is compatible with the expressions derived.  

% -------------------------------------- %
% ---------- Entropy Steering ---------- %
% -------------------------------------- %

\subsection{Relative Entropy Steering}

In this section, we show that the time rate of change in the relative entropy can be approximately bounded as 
\begin{flalign}
    \frac{d}{dt}\mathcal{R}\!\left(\pi_t\middle\|\hat{\pi}_t\right)
&\lesssim \dot{\mathcal{R}}^{\text{ub}}(\pi_t\| \hat{\pi}_t)\nonumber\\
&= 
\frac{1}{2}\mathbb{E}_{\hat{\pi}_t}
\left[
\left\|
D^{-1/2}\left(\boldsymbol{\phi}_t-\hat{\boldsymbol{f}}_t\right)
\right\|_2^2
\right],\label{eqn:entropy_rate_ub}
\end{flalign}
and because Eq. \ref{eqn:entropy_rate_ub} is nonnegative,  
\begin{flalign}
    \mathcal{R}(\pi_t\|\hat{\pi}_t) &\lesssim \mathcal{R}^{\text{ub}}(\pi_t\|\hat{\pi}_t)\nonumber\\
    &= \mathcal{R}(\pi_0\|\hat{\pi}_0) + \int_0^t \dot{\mathcal{R}}^{\text{ub}}(\pi_\tau\|\hat{\pi}_\tau) d\tau
\end{flalign}
Above, the matrix $D = GQG^\top$. While not required, we may assume that $\pi_0$ is known, and thus $\mathcal{R}(\pi_0\| \hat{\pi}_0)$ is computable. For simplicity we take $\pi_0 = \hat{\pi}_0$, therefore, $\mathcal{R}(\pi \|\hat{\pi}_0) = 0$. In cases where the initial true distribution is unknown, it may be desirable to use a conservative value for $\mathcal{R}(\pi_0\|\hat{\pi}_0)$, potentially informed by domain knowledge available to the user. 

The derivation of the inequality in Eq. \ref{eqn:entropy_rate_ub} is introduced in our previous work \cite{Wolf_2026}, and is similar to that originally provided in \cite{Mou_2022}. We discuss it here again for reader clarity and completeness. First, notice that 
\begin{equation}
    \frac{\partial}{\partial t}
    \left(
    \pi_t \ln \frac{\pi_t}{\hat{\pi}_t}
    \right)
    =
    \pi_t
    \left(
    \frac{\partial \ln \pi_t}{\partial t}
    \left(
    \ln \pi_t + 1 - \ln \hat{\pi}_t
    \right)
    -
    \frac{\partial \ln \hat{\pi}_t}{\partial t}
    \right).
\end{equation}
The left-hand side above is the time rate of change of the integrand in Eq.~\ref{eqn:kld}, i.e., the integrand of the \ac{KLD} definition, and under mild assumptions, it can be shown that
\begin{equation}
    \frac{d}{dt}\mathcal{R}\!\left(\pi_t\middle\|\hat{\pi}_t\right)
    =
    \int_{\mathcal{X}}
    \frac{\partial \pi_t}{\partial t}
    \left(
    \ln \pi_t+1-\ln \hat{\pi}_t
    \right)
    \,d\boldsymbol{x}_t
    -
    \int_{\mathcal{X}}
    \frac{\partial \hat{\pi}_t}{\partial t}
    \frac{\pi_t}{\hat{\pi}_t}
    \,d\boldsymbol{x}_t.
\end{equation}
For a general probability distribution $q_t$, the Fokker-Planck equation describes its time evolution,
\begin{equation}
    \frac{\partial q_t}{\partial t}
    =
    -\nabla_{\boldsymbol{x}_t}\cdot\left(q_t\boldsymbol{\phi}_t\right)
    +\frac{1}{2}\nabla_{\boldsymbol{x}_t}\cdot\left(D\nabla_{\boldsymbol{x}_t}\pi_t\right).
\end{equation}
Substituting this form for the time derivatives of $\pi_t$ and $\hat{\pi}_t$ and reducing, we obtain
\begin{flalign}
    \frac{d}{dt}\mathcal{R}\!\left(\pi_t\middle\|\hat{\pi}_t\right)
    &=
    \int_{\mathcal{X}}
    \pi_t\left(\boldsymbol{\phi}_t-\boldsymbol{\hat{f}}_t\right)\cdot
    \left(
    \nabla_{\boldsymbol{x}_t}\ln\pi_t
    -
    \nabla_{\boldsymbol{x}_t}\ln\hat{\pi}_t
    \right)
    \,d\boldsymbol{x}_t\\
    &-
    \frac{1}{2}
    \int_{\mathcal{X}}
    \left\|
    D^{1/2}
    \left(
    \nabla_{\boldsymbol{x}_t}\ln\pi_t
    -
    \nabla_{\boldsymbol{x}_t}\hat{\pi}_t
    \right)
    \right\|_2^2
    \,d\boldsymbol{x}_t.
\end{flalign}
Next, invoking Young's inequality, i.e., 
\begin{equation}
    \boldsymbol{a} \cdot \boldsymbol{b} \leq \frac{1}{2}(\|\boldsymbol{a}\|_2^2 + \|\boldsymbol{b}\|_2^2),
\end{equation}
we obtain
\begin{flalign}
    \frac{d}{dt}\mathcal{R}\!\left(\pi_t\middle\|\hat{\pi}_t\right)
    &\leq
    \frac{1}{2}\int_{\mathcal{X}}
    \pi_t
    \left\|
    D^{-1/2}\left(\boldsymbol{\phi}_t-\hat{\boldsymbol{f}}_t\right)
    \right\|_2^2
    \,d\boldsymbol{x}_t\\
    &=
    \frac{1}{2}\mathbb{E}_{\pi_t}
    \left[
    \left\|
    D^{-1/2}\left(\boldsymbol{\phi}_t-\hat{\boldsymbol{f}}_t\right)
    \right\|_2^2
    \right].
\end{flalign}
The reader should note that the second expression above still requires an expectation evaluated under the true distribution $\pi_t$. A formal treatment of this problem is discussed in our previous work \cite{Wolf_2026}. Here, however, we posit that, provided $\mathcal{R}(\pi_t\| \hat{\pi}_t)$ is constrained to be small, Eq. \ref{eqn:entropy_rate_ub} provides a useful approximate upper bound for the distance time rate of change. That is, we can evaluate this expectation under our approximate distribution.    

\subsubsection{Simplification with a Gaussian Reference:}
From here, we assume that errors in the dynamics function $\hat{\boldsymbol{f}}_t$ used in a guidance algorithm are solely due to structural approximations (i.e., linearization). In other words, the user has accurate knowledge of the nonlinear dynamics of the system so that $\boldsymbol{\phi}_t = \boldsymbol{f}_t$. Future work will investigate cases where this may not be true. Furthermore, let us assume that linearization errors are dominated by the second-order truncated term so that 
\begin{flalign}
    \boldsymbol{f}_t - \boldsymbol{\hat{f}}_t&=\frac{1}{2}\delta\boldsymbol{x}_t^{\top}\nabla_{\boldsymbol{x}_t}^{2}\boldsymbol{f}_t(\boldsymbol{x}^*_t)\delta\boldsymbol{x}_t+\mathcal{O}\!\left(\left\|\delta\boldsymbol{x}_t\right\|^{3}\right)\\
    &\simeq\frac{1}{2}\left[\delta\boldsymbol{x}_t^{\top}H_t^{1}\delta\boldsymbol{x}_t,\ldots,\delta\boldsymbol{x}_t^{\top}H_t^{n}\delta\boldsymbol{x}_t\right]^{\top}.
\end{flalign}
Above, each $H_t^i$ is the $i^{\text{th}}$ frontal slice of the second-order dynamics derivative tensor evaluated at the reference state $\boldsymbol{x}_t^*$. If we assume that the acceleration process noise is spherical, i.e., $D = \sigma^2 I_{n\times n}$, substituting into Eq. \ref{eqn:entropy_rate_ub},
\begin{equation}
\mathbb{E}_{\hat{\pi}_t}\!\left[\left\|D^{-1/2}(\boldsymbol{f}_t-\boldsymbol{\hat{f}}_t)\right\|_2^2\right]
    =\frac{1}{4\sigma^2}\sum_{i=1}^{n}\mathbb{E}_{\hat{\pi}_t}\!\left[\left(\delta\boldsymbol{x}_t^{\top}H_t^i\delta\boldsymbol{x}_t\right)^2\right].
\end{equation}
Now, for any square matrix, $S \in \mathbb{R}^{n\times n}$, and random variable $\boldsymbol{\xi} \sim \mathcal{N}(\boldsymbol{\xi}; 0_{n\times 1}, \Sigma)$, it true that \cite{Magnus_1978} 
\begin{equation}
    \mathbb{E}_{\mathcal{N}(\boldsymbol{\xi}; 0_{n\times 1}, \Sigma)}\!\left[\left(\boldsymbol{\xi}^{\top}S\boldsymbol{\xi}\right)^2\right]
=
\operatorname{tr}(S\Sigma)^2
+
2\operatorname{tr}\!\left((S\Sigma)^2\right).
\end{equation}
Therefore, 
\begin{equation}
    \mathbb{E}_{\hat{\pi}_t}\!\left[\left\|D^{-1/2}(\boldsymbol{f}_t-\boldsymbol{\hat{f}}_t)\right\|_2^2\right]=\frac{1}{4\sigma^2}\sum_{i=1}^{n}\left[\operatorname{tr}\!\left(H_t^iP_t\right)^2+2\operatorname{tr}\!\left(\left(H_t^iP_t\right)^2\right)\right],
\end{equation}
and
\begin{flalign}
    \frac{d}{dt}\mathcal{R}(\pi_t\| \hat{\pi}_t) &\lesssim \dot{\mathcal{R}}^{\text{ub}}(\pi_t\|\hat{\pi}_t) \nonumber \\
    & =
    \frac{1}{8\sigma^2}\sum_{i=1}^{n}\left[\operatorname{tr}\!\left(H_t^iP_t\right)^2+2\operatorname{tr}\!\left(\left(H_t^iP_t\right)^2\right)\right].\label{eqn:final_upper_bound}
\end{flalign}

The reader should notice that Eq. \ref{eqn:final_upper_bound} is functionally dependent on $P_t$ -- a decision variable in the covariance steering algorithm. It stands to reason then that $\mathcal{R}^{\text{ub}}(\cdot)$ is controllable under this guidance framework, a feature we exploit in the following section. Another salient feature is the inverse-squared relation between the distance rate upper bound and the diffusion magnitude, $\sigma$. This suggests that, in the large-diffusion regime, i.e., large process noise, the Gaussian surrogate more accurately captures the evolution of the true nonlinear distribution. In the following section, we introduce an approach for steering this relative entropy upper bound using the \ac{SCvx} algorithm  \cite{Mao_2018}. 

% -------------------------------- %
% ------------- SCVX ------------- %
% -------------------------------- %

\subsection{Implementation with \ac{SCvx}}

\subsubsection{Standard Covariance Steering:}
To begin, we briefly revisit the standard covariance steering problem. Written in continuous time, we seek a linear feedback policy of the form
\begin{equation}
\tilde{\boldsymbol{u}}_t = K_t \tilde{ \boldsymbol{x}}_t.
\end{equation}
where $K_t$ is the control gain matrix and $\tilde{\boldsymbol{u}}_t$ is the control feedback so that $\boldsymbol{u}_t = \boldsymbol{u}^*_t + \tilde{\boldsymbol{u}}_t$. The variable $\boldsymbol{u}^*_t$ is the nominal open-loop control. The linearized state deviation flow can then be expressed as  
\begin{flalign}
\dot{\tilde{\boldsymbol{x}}}_t &= A_t \tilde{\boldsymbol{x}}_t + B_t \tilde{ \boldsymbol{u}}_t + G \boldsymbol{w}_t\\
&= (A_t + B_t K_t)\tilde{\boldsymbol{x}}_t + G \boldsymbol{w}_t .
\end{flalign}
Where $A_t$ and $B_t$ are the Jacobian matrices of the dynamics vector $\boldsymbol{f}_t$ taken with respect to the state and and control, and evaluated at $\boldsymbol{x}_t^*$ and $\boldsymbol{u}_t^*$, respectively. For computer implementation, we use a \ac{FOH} interpolation scheme to discretize this expression so that the discrete time deviation dynamics are
\begin{equation}
\tilde{\boldsymbol{x}}_{k+1} = \left(I - B_k^{-}K_{k+1}\right)^{-1}
\left[(A_k + B_k^{-}K_k)\tilde{\boldsymbol{x}}_k + \boldsymbol{w}_k\right] .
\end{equation}
The covariance of the state deviations, $P_k = \mathbb{E}_{\hat{\pi}_t}[\tilde{\boldsymbol{x}}_k\tilde{\boldsymbol{x}}_k^\top]$, then evolves in discrete time accordingly:
\begin{equation}
\begin{aligned}
P_{k+1}
={}& A_k P_k A_k^{\top}
+ A_k U_k^{\top} B_k^{-\top}
+ B_k^{-} U_k A_k^{\top}
+ B_k^{-} Y_k B_k^{-\top} \\
&+ U_{k+1}^{\top} B_k^{+\top}
+ B_k^{+} U_{k+1}
- B_k^{+} Y_{k+1} B_k^{+\top}
+ D_k, \hspace{1ex} \text{for } k = 1:N
\end{aligned}
\label{eqn:const_1}
\end{equation}
subject to 
\begin{equation}
\begin{bmatrix}
P_k & U_k^{\top} \\
U_k & Y_k
\end{bmatrix}
\succeq 0\hspace{1ex} \text{for } k = 1:N,
\label{eqn:const_2}
\end{equation}
where $N$ is the number of discretization nodes. The constituent matrices above are as follows:
\begin{equation}
A_k = \Phi(t_{k+1}, t_k),
\end{equation}

\begin{equation}
B_k^{-} = A_k \int_{t_k}^{t_{k+1}}
\frac{t_{k+1}-\tau}{t_{k+1}-t_k}
\Phi^{-1}(\tau, t_k)\, B \, d\tau,
\end{equation}

\begin{equation}
B_k^{+} = A_k \int_{t_k}^{t_{k+1}}
\frac{\tau-t_k}{t_{k+1}-t_k}
\Phi^{-1}(\tau, t_k)\, B \, d\tau,
\end{equation}

\begin{equation}
D_k = \int_{t_k}^{t_{k+1}}
\Phi(t_{k+1}, \tau)\, GG^{\top}\, \Phi^{\top}(t_{k+1}, \tau)\, d\tau,
\end{equation}

\begin{equation}
U_k = K_k P_k,
\end{equation}

\begin{equation}
Y_k = U_k P_k^{-1} U_k^{\top}.
\end{equation}
The standard covariance steering problem can then be stated as a convex semidefinite program,
\begin{subequations}
    \begin{flalign}
        &\hspace{-10ex}\min_{\boldsymbol{X}} J(\boldsymbol{X})\\
        &\hspace{-5ex}\text{s.t. Eq. \ref{eqn:const_1}, \ref{eqn:const_2}},\\
        &\hspace{-5ex} P_1 = P^*_1, P_N \preceq P^*_N.
    \end{flalign}
\end{subequations}
Above, $P_1^*$ and $P_N^*$ are the initial and terminal target covariance matrices. The set $\boldsymbol{X} \coloneqq \{P_k, Y_k, U_k \}_{k = 1}^N$, and the cost function minimizes the cumulative feedback control variance:
\begin{equation}
J
=
\sum_{k=1}^{N-1}
\frac{\Delta t_k}{2}
\left(
\operatorname{tr}(Y_k)
+
\operatorname{tr}(Y_{k+1})
\right),
\end{equation}
where $\Delta t_k = t_{k + 1} - t_k$.

\subsubsection{Convexification of Relative Entropy Constraints:}

We treat the the relative entropy upper bound derived previously as a decision variable in \ac{SCvx}. For a detailed review of the \ac{SCvx} algorithm, we encourage the reader to refer to Ref. \citenum{Malyuta_2022}. 

The rate upper bound evaluated at discrete time $t_k$ is
\begin{equation}
    \dot{\mathcal{R}}^{\text{ub}}_k(\pi_t\| \hat{\pi}_t) = \frac{1}{8\sigma^2}\sum_{i = 1}^n\left[\operatorname{tr}(H_k^iP_k)^2 + 2\operatorname{tr}\left((H_k^iP_k)^2\right) \right],\label{eqn:entropy_rate_ub_new}
\end{equation}
and a first-order Euler step integrates this equation so that
\begin{equation}
    \mathcal{R}^{\text{ub}}_{k+ 1} = \mathcal{R}^{\text{ub}}_k + \Delta t_k\dot{\mathcal{R}}^{\text{ub}}_k.
\end{equation}
The quantity $\dot{\mathcal{R}}_k^{\text{ub}}$ is nonlinear in the decision variable $P_k$. To be compatible with each convex programming step, we linearize the dynamics so that 
\begin{equation}
    \mathcal{R}_{k + 1}^{\text{ub}} = \mathcal{R}_{k}^{\text{ub}} + \Delta t_k\left(\dot{\bar{\mathcal{R}}}^{\text{ub}}_k +   \operatorname{tr}\left[\overline{\nabla_{P_k}\dot{\mathcal{R}}^{\text{ub}}_k} \cdot (P_k - \bar{P}_k)\right]  \right) + \omega_k, \hspace{3ex} k = 1:N - 1, 
    \label{eqn:const_3}
\end{equation}
where the matrix 
\begin{equation}
    \overline{\nabla_{P_k}\dot{\mathcal{R}}^{\text{ub}}_k} = \frac{1}{4 \sigma^2} \sum_{i = 1}^n \left[\operatorname{tr}(H_k^i \bar{P}_k)H_k^i + 2H_k^i \bar{P}_k H_k^i \right],
\end{equation}
and the reference bound rate is 
\begin{flalign}
    \dot{\bar{\mathcal{R}}}_k^{\text{ub}} = \frac{1}{8\sigma^2}\sum_{i = 1}^n \left[\operatorname{tr}(H_k^i \bar{P}_k)^2 + 2\operatorname{tr}\left((H_k^i \bar{P}_k)^2 \right) \right].
\end{flalign}
Above, $\bar{P}_k$ is the covariance provided by the solution of the previous convex step. The scalar decision variable $\omega_k$ is a slack variable that must be driven to zero through successive iterations to ensure dynamical consistency of the integrated upper bound. 

\subsubsection{Convexification of the \ac{MSE} Variational Upper Bound:}

Like $\mathcal{R}_k^{\text{ub}}$, we can treat the upper bound for an expected quantities of interest as a decision variable within \ac{SCvx}. The following shows the approach taken for the \ac{MSE}, but a similar procedure can be used for the risk probability bound derived above. Define the discrete time \ac{MSE} bound with the variable $\epsilon_k^{\text{ub}}$ so that
\begin{equation}
    \mathbb{E}_{\pi_k}[\tilde{\boldsymbol{x}}_k^\top \tilde{\boldsymbol{x}}_k] \leq \epsilon^{\text{ub}}_k = \frac{1}{\lambda_k} \left\{-\frac{1}{2}\operatorname{logdet}(I - 2\lambda_kP_k) + \mathcal{R}_k^{\text{ub}}  \right\}\label{eqn:MSE_upper_bound}
\end{equation}
The right-hand side of Eq. \ref{eqn:MSE_upper_bound} is convexified so that at each convex iteration we satisfy
\begin{equation}
    \epsilon^{\text{ub}}_k = \bar{\epsilon}_k^{\text{ub}} + \overline{\frac{\partial \epsilon_k^{\text{ub}}}{ \partial\lambda_k}} \cdot(\lambda_k - \bar{\lambda}_k) + \operatorname{tr}\left[\overline{\nabla_{P_k}\epsilon_k^{\text{ub}}}\cdot(P_k - \bar{P}_k) \right] + \overline{\frac{\partial \epsilon_k^{\text{ub}}}{\partial R_k^{\text{ub}}}}\cdot (\mathcal{R}_k^{\text{ub}} - \bar{\mathcal{R}}_k^{ub}) + \nu_k.
    \label{eqn:const_4}
\end{equation}
The reference quantity is provided as 
\begin{equation}
    \bar{\epsilon}_k^{\text{ub}} = \frac{1}{\bar{\lambda}_k}\left\{ -\frac{1}{2}\operatorname{logdet}(I - 2 \bar{\lambda}_k \bar{P}_k) + \bar{\mathcal{R}}_k^{\text{ub}} \right\},
\end{equation}
and the derivatives in Eq. \ref{eqn:const_4}, evaluated at the previous iteration's solution $\{\bar{\lambda}_k, \bar{P}_k, \bar{\mathcal{R}}_k^{\text{ub}}\}_{k = 1}^N$, are
\begin{equation}
    \overline{\frac{\partial \epsilon_k^{\text{ub}}}{\partial \lambda_k}} = \frac{1}{\bar{\lambda}_k}\operatorname{tr}\left[(I - 2\bar{\lambda}_k \bar{P}_k)^{-1}\bar{P}_k \right] - \frac{1}{\bar{\lambda}_k^2}\left[\bar{\mathcal{R}}_k^{\text{ub}} - \frac{1}{2} \operatorname{logdet}(I - 2\bar{\lambda}_k \bar{P_k}) \right],
\end{equation}
\begin{equation}
    \overline{\nabla_{P_k} \epsilon_k^{\text{ub}}} = (I - 2\bar{\lambda}_k \bar{P}_k)^{-1},
\end{equation}
and,
\begin{equation}
    \overline{\frac{\partial \epsilon_k^{\text{ub}}}{\partial \mathcal{R}_k^{\text{ub}}}} = \frac{1}{\bar{\lambda}_k}.
\end{equation}
The decision variables $\{\nu_k\}_{k = 1}^N$ are slack terms that must be driven to zero in order to satisfy Eq. \ref{eqn:MSE_upper_bound}.

Since Eq.~\ref{eqn:MSE_upper_bound} is convex in the decision variable 
\(\lambda_k\), any feasible stationary point is a global minimizer. Therefore,
the optimal value \(\lambda_k^\star\) can be obtained by enforcing the 
first-order optimality condition
\begin{equation}
    \frac{\partial \epsilon_k^{\text{ub}}}{\partial \lambda_k} = 0 \Longleftrightarrow \underbrace{\lambda_k\operatorname{tr}\left[(I - 2\lambda_k P_k)^{-1}P_k \right] + \frac{1}{2}\operatorname{logdet}(I - 2\lambda_kP_k) - \mathcal{R}_k^{\text{ub}}}_{\psi_k(\cdot)} = 0 \label{eqn:FON_cond}
\end{equation}
Each convex program iteration satisfies a linear expansion of the optimality condition 
\begin{flalign}
    &\bar{\psi_k} + \overline{\frac{\partial \psi_k}{\partial \lambda_k}} \cdot (\lambda_k - \bar{\lambda}_k) + \operatorname{tr}\left[\overline{\nabla_{P_k} \psi_k} \cdot (P_k - \bar{P}_k)\right] + \overline{\frac{\partial \psi_k}{\partial \mathcal{R}_k^{\text{ub}}}} \cdot (\mathcal{R}_k^{\text{ub}} - \bar{\mathcal{R}}_k^{\text{ub}}) + \nu'_k = 0,\\
    &\hspace{5ex} \text{for } k = 1:N,
    \label{eqn:const_5}
\end{flalign}
where the derivatives evaluated at the previous solution are 
\begin{flalign}
    \overline{\frac{\partial \psi_k}{\partial \lambda_k}} = &\operatorname{tr}\left[ (I - 2\bar{\lambda}_k\bar{P}_k)^{-1}\bar{P}_k\right] + 2\lambda_k \operatorname{tr}\left[(I - 2\bar{\lambda}_k\bar{P}_k)^{-2} \bar{P}_k^2\right]\\
    &- \operatorname{tr}\left[(I - \bar{\lambda}_k\bar{P}_k)^{-1}\bar{P}_k \right],
\end{flalign}
\begin{equation}
\overline{\nabla_{P_k} \psi_k} = \bar{\lambda}_k\left( (I - 2\bar{\lambda}_k\bar{P_k})^{-2} - (I - 2\bar{\lambda}_k \bar{P}_k)^{-1} \right),
\end{equation}
and, 
\begin{equation}
    \overline{\frac{\partial\psi_k}{ \partial \mathcal{R}_k^{ub}}} = -1.
\end{equation}
Once more, the set of slack variables $\{\nu_k' \}_{k = 1}^N$ must be driven to zero through successive iterations to satisfy the first-order optimality condition. 

\subsubsection{Form of the Convex Subproblem:} 
In this study, we consider two related problem formulations. The first constrains the upper bound on the relative entropy to remain below a prescribed threshold, while the second constrains the derived upper bound on the \ac{MSE} to remain below a prescribed threshold. Combining the preceding results, each convex step of the first problem is formulated as

\begin{subequations}
\begin{flalign}
    &\hspace{-5ex}\min_{\boldsymbol{Z}} \mathcal{L}(\boldsymbol{Z}; \bar{\boldsymbol{Z})}\\
    &\hspace{-0ex}\text{s.t. Eq. \ref{eqn:const_1}, \ref{eqn:const_2}, \ref{eqn:const_3}},\\
    &\hspace{-0ex} P_1 = P_1^*,\hspace{1ex} P_N \preceq P_N^*\\
    &\hspace{-0ex} \mathcal{R}_1^{\text{ub}} = 0, \hspace{1ex} \mathcal{R}_N^{\text{ub}} \leq {\mathcal{R}_N^{\text{ub}}}^*\\
    &\hspace{-0ex}\|P_k - \bar{P}_k\|_{F}  \leq \eta, \hspace{1ex} \text{for } k = 1:N.
\end{flalign}
\label{eqn:subproblem_form_1}
\end{subequations}
where the variable set $\boldsymbol{Z} \coloneqq \{P_k, Y_k, U_k, \mathcal{R}_k^{\text{ub}}, \omega_k \}_{k = 1}^N$. The subproblem cost function is given as 
\begin{equation}
    \mathcal{L}(\boldsymbol{Z}; \bar{\boldsymbol{Z}}) = \sum_{k = 1}^{N - 1}\left[\frac{\Delta t_k}{2}\left(\operatorname{tr}(Y_k) + \operatorname{tr}(Y_{k + 1}) \right)\right] + \gamma \sum_{i = 1}^N |\omega_k|.
\end{equation}
The trust region radius $\eta$ is updated after each iteration according to the protocol prescribed in \cite{Malyuta_2022}. The positive scalar $\gamma$ is a tuning variable, and must be chosen to be sufficiently large to drive the slack variables to zero. 

Putting the pieces together for the second formulation, each convex step within \ac{SCvx} solves
\begin{subequations}
\begin{flalign}
    &\hspace{-5ex}\min_{\boldsymbol{Z}} \mathcal{L}(\boldsymbol{Z}; \bar{\boldsymbol{Z})}\\
    &\hspace{-0ex}\text{s.t. Eq. \ref{eqn:const_1}, \ref{eqn:const_2}, \ref{eqn:const_3}, \ref{eqn:const_4}, \ref{eqn:const_5}},\\
    &\hspace{-0ex} \epsilon_{k}^{\text{ub}} \leq \text{MSE}^*, \hspace{1ex} \text{for } k = 1:N,\\
    &\hspace{-0ex} P_1 = P_1^*\\
    &\hspace{-0ex} \mathcal{R}_1^{\text{ub}} = 0\\
    &\hspace{-0ex}\|P_k - \bar{P}_k\|_{F} + |\lambda_k - \bar{\lambda}_k| + |\mathcal{R}_k^{\text{ub}} - \bar{\mathcal{R}}_k^{\text{ub}}| \leq \eta, \hspace{1ex} \text{for } k = 1:N.
\end{flalign}
\label{eqn:subproblem_form_2}
\end{subequations}
The complete set of decision variables is provided by 
\begin{equation}
    \boldsymbol{Z} \coloneqq \{P_k, Y_k, U_k, \lambda_k, \mathcal{R}_k^{\text{ub}}, \epsilon_k^{\text{ub}}, \omega_k, \nu_k, \nu'_k\}_{k = 1}^N,
\end{equation}
and the problem's cost function is 
\begin{equation}
    \mathcal{L}(\boldsymbol{Z}; \bar{\boldsymbol{Z}}) = \sum_{k = 1}^{N - 1}\left[\frac{\Delta t_k}{2}\left(\operatorname{tr}(Y_k) + \operatorname{tr}(Y_{k + 1}) \right)\right] + \gamma \sum_{i = 1}^N \Big[|\omega_k| + |\nu_k| + |\nu'_k|\Big].
\end{equation}
%

% ------------------------
% ------- Results --------
% ------------------------

\section{Results}

\subsection{Experiment Setup}

To assess the proposed methods, we use a numerical test case involving a transfer trajectory between two \acp{NRHO} in the \ac{CRTBP}, with the Earth and moon as the primary and secondary central bodies, respectively. The equations of motion are given by 
\begin{flalign}
    \ddot{x} &= 2\dot{y} + x 
    - (1-\mu)\frac{x+\mu}{r_1^3}
    - \mu\frac{x+\mu-1}{r_2^3},\nonumber \\
    \ddot{y} &= -2\dot{x} + y
    - (1-\mu)\frac{y}{r_1^3}
    - \mu\frac{y}{r_2^3}, \nonumber \\
    \ddot{z} &= -(1-\mu)\frac{z}{r_1^3}
    - \mu\frac{z}{r_2^3}, 
\end{flalign}
where 
\begin{flalign}
    r_1^2 &= (x + \mu)^2 + y^2 + z^2, \nonumber \\
    r_2^2 &= (x + \mu -1)^2 + y^2 + z^2. 
\end{flalign}
The constant $\mu = \frac{m_2}{m_1 + m_2}$ with $m_1$ and $m_2$ the masses of the primary and secondary. The test case we select is well set to demonstrate the advantages of the proposed methods due to existing large nonlinearities during close encounters of the spacecraft with the moon. In these regions, standard linearized covariance steering guidance policies struggle to generate corrections that are consistent with the nonlinear dynamics. 

\begin{table}[htb!]
    \centering
    \caption{Reference transfer initial and terminal conditions.}
    \label{tab:reference_transfer}
    \begin{tabular}{c c c}
        \hline
        State & $\boldsymbol{x}_{1}^*$ & $\boldsymbol{x}_{N}^*$ \\
        \hline
        $x_0$ [DU]        & $1.0682401667504136$             & $1.1221016015401843$ \\
        $y_0$ [DU]        & $0.0$                            & $0.0$ \\
        $z_0$ [DU]       & $-2.0127217384604126\times 10^{-1}$ & $-1.8496190168752416\times 10^{-1}$ \\
        $\dot{x}_0$ [DU/TU]  & $0.0$                            & $0.0$ \\
        $\dot{y}_0$ [DU/TU]  & $-1.8366091421944192\times 10^{-1}$ & $-2.2509936715835580\times 10^{-1}$ \\
        $\dot{z}_0$ [DU/TU] & $0.0$                            & $0.0$ \\
        \hline
        $T$ [TU]          & $2.1611242376925008$             & $2.9227417192680303$ \\
        \hline
    \end{tabular}
\end{table}

The nominal low-thrust transfer trajectory is generated with the \ac{SCvx} algorithm. The initial and terminal state conditions, along with the periods of the respective orbits of those states, are provided in Table \ref{tab:reference_transfer}. Units are normalized by a characteristic distance and time given as $\text{DU} = 389703.0$ km and $\text{TU} = 382981.0$ s. For the Earth-moon system, the mass ratio $\mu = 1.215058560962404\times10^{-2}$. The spacecraft has a maximum thrust acceleration of $90 \hspace{1ex}\mu\text{m}/\text{s}^2$. 

% Nominal trajectory 
\begin{figure}[htb!]
    \centering
    \includegraphics[width = 0.8\textwidth]{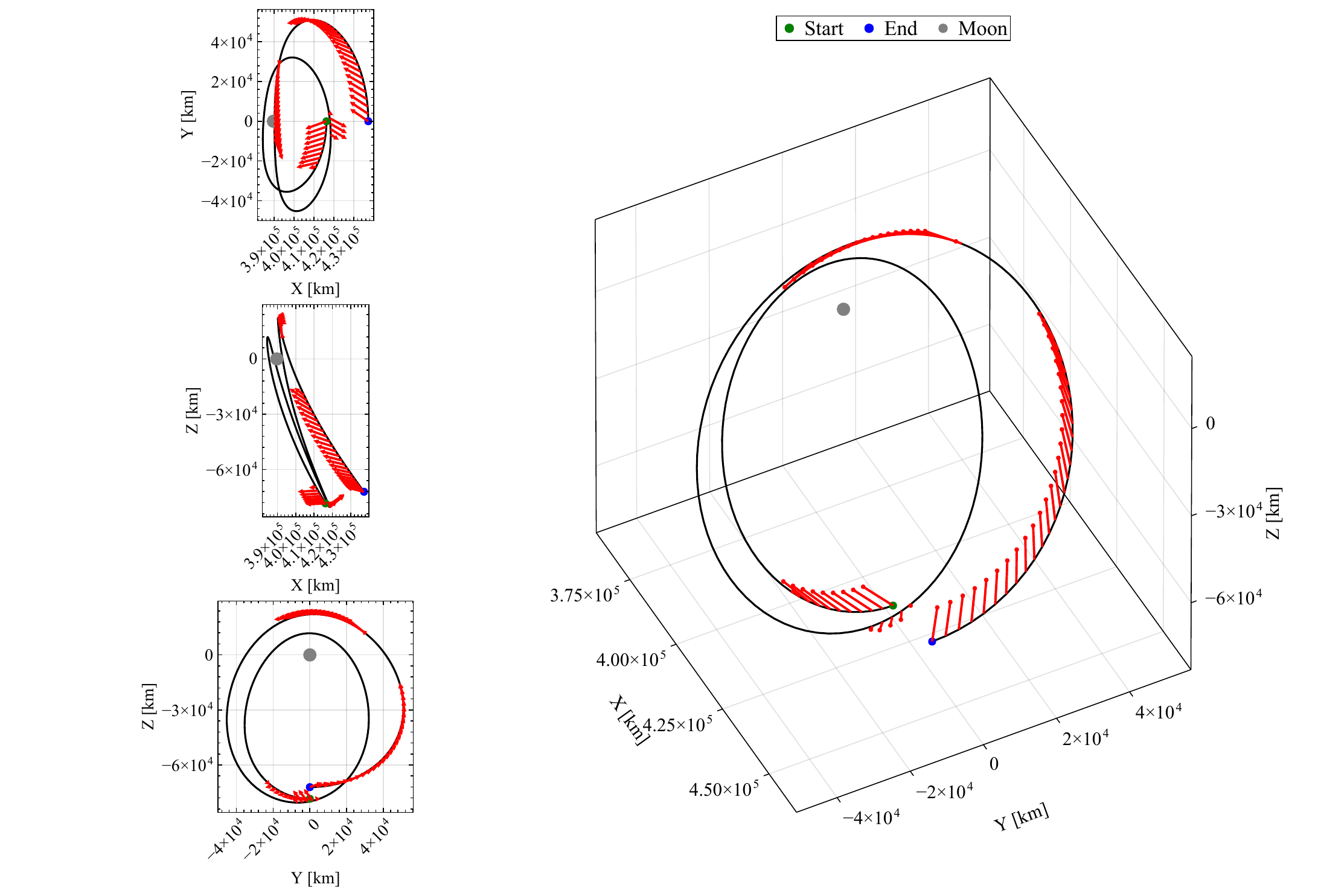}
    \caption{Nominal transfer between two \acp{NRHO} shown in the Earth-moon rotating frame.}
    \label{fig:nominal_trajectory}
\end{figure}

\begin{figure}[htb!]
    \centering
    \includegraphics[width = 0.7\textwidth]{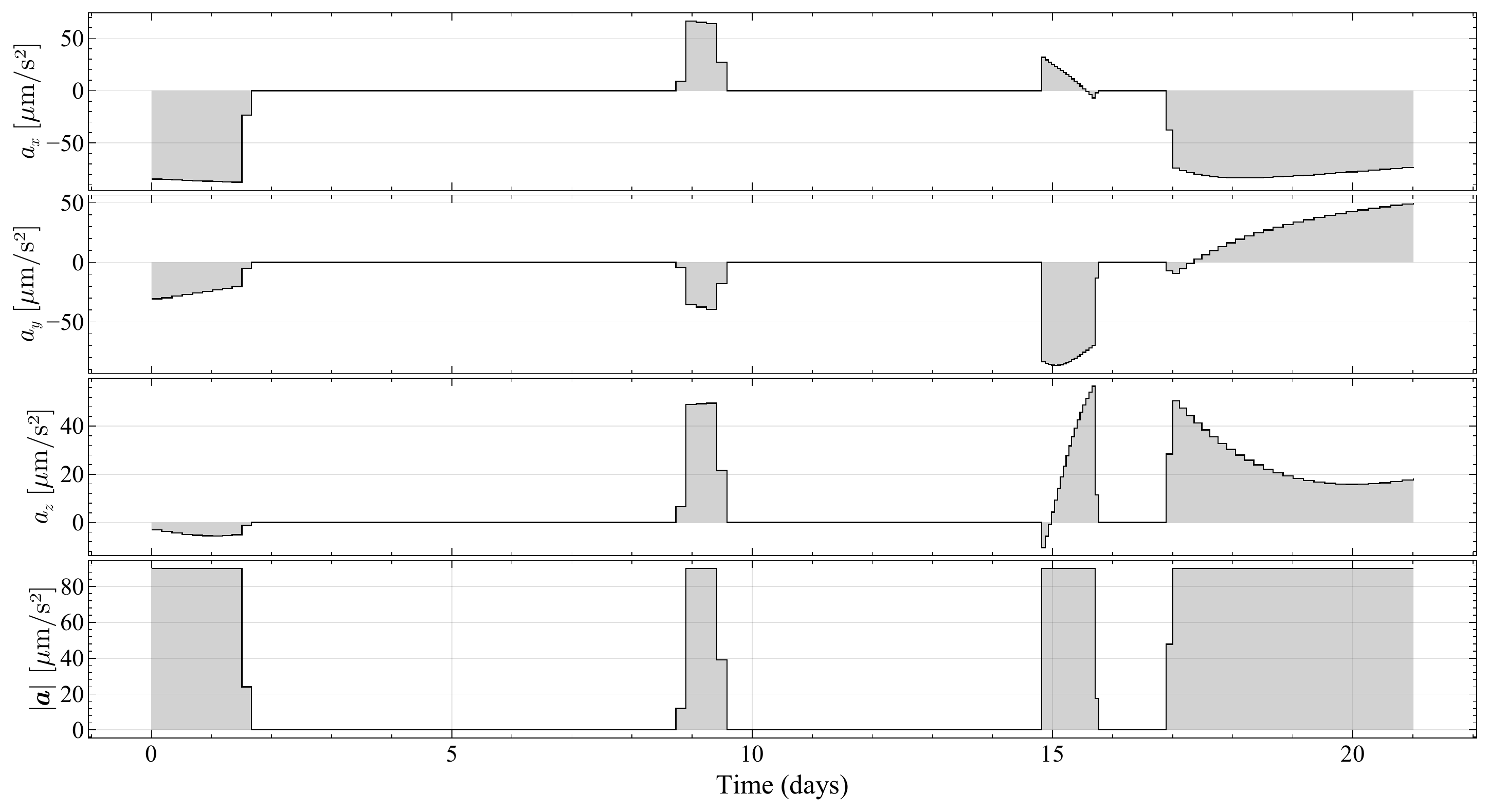}
    \caption{Thrust profile for the nominal transfer.}
    \label{fig:nominal_thrust}
\end{figure}

Figure \ref{fig:nominal_trajectory} shows the nominal transfer in the Earth-moon rotating frame. The right-hand side is a 3D view of the trajectory, along with red quivers indicating the thrust magnitude and direction. The left-hand side shows the projection of this transfer onto each of the coordinate planes. There is a close approach with the moon at about 3 days after the starting epoch. In Figure \ref{fig:nominal_thrust} we show the thrust profile for the nominal transfer trajectory. The top three panels show $x$, $y$, and $z$ directions, and the bottom shows the thrust magnitude. The total duration of the scenario is 4.74 TU or about $21$ days.  

% Nominal thrust curves 

\subsection{Test Case \#1: Relative Entropy Targeting}

Presently, we intend to isolate the problem of constraining the relative entropy between the true distribution and its surrogate using an iterative covariance steering procedure. To that end, the subproblem stated in Eq. \ref{eqn:subproblem_form_1} is solved successively using \ac{SCvx}. The conditions for this test case are stated in Table \ref{tab:conditions_case_12}. 
\begin{table}[htb!]
    \centering
    \caption{Prescribed conditions for test case \#1 and \#2.}
    \label{tab:conditions_case_12}
    \begin{tabular}{c c c}
        \hline
        Value & Case \#1 & Case \#2 \\
        \hline
         Initial RMS Position (Per Axis) [km]  & 400.0             & 400.0 \\
        Initial RMS Velocity (Per Axis) [km/s] & $4.0\times10^{-3}$  & $4.0\times10^{-3}$ \\
        Terminal RMS Position (Per Axis) [km]  & $400.0$ & -- \\
        Terminal RMS Velocity (Per Axis) [km/s] & $4\times10^{-3}$ & -- \\
        Process Noise RMS (Per Axis), $\sigma$ [km/$\text{s}^{3/2}$] & $1\times10^{-6}$ & $1\times10^{-6}$ \\
        Relative Entropy Bound, ${\mathcal{R}^{\text{ub}}_N}^*$ & 100.0, 10.0, 1.0, 0.1 & -- \\
        Terminal MSE, $\operatorname{tr}(P_N)$ [$\text{nd}^2$] & --  & $10^{-4}$\\
        \hline
    \end{tabular}
\end{table}

% Entropy plot
\begin{figure}[htb!]
    \centering
    \includegraphics[width = 0.7\textwidth]{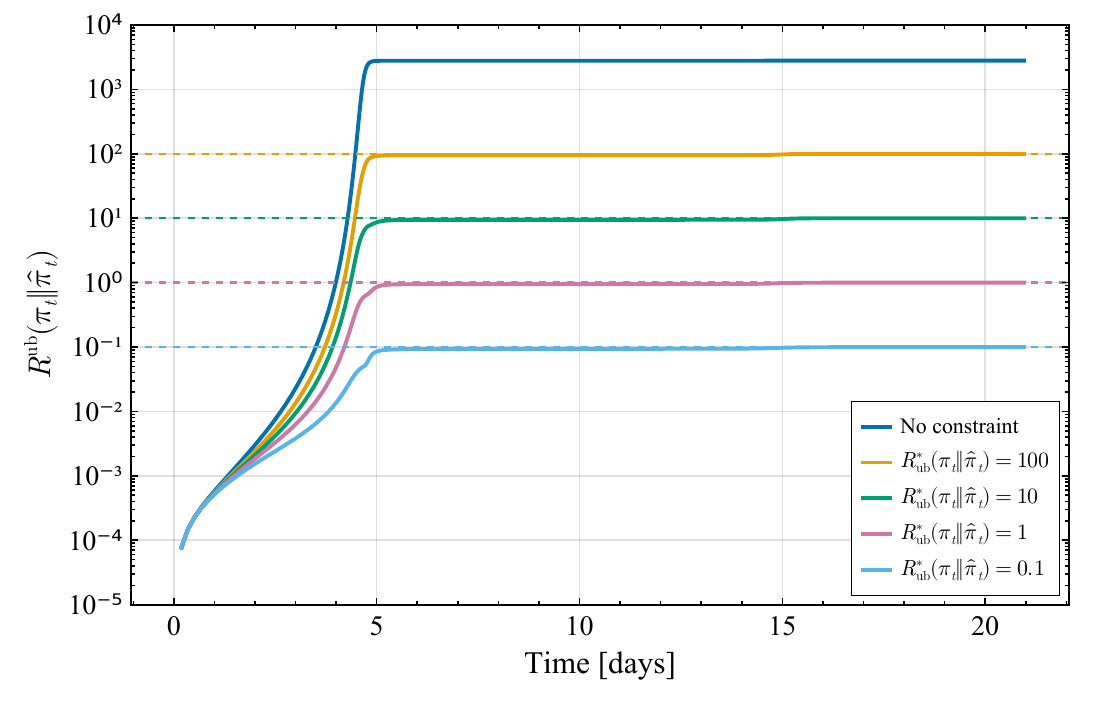}
    \caption{Relative entropy upper bound history subject to bound constraints. The solid dark blue line is the natural evolution with standard covariance steering.}
    \label{fig:tc1_entropy}
\end{figure}

% Trace Plot
\begin{figure}[htb!]
    \centering
    \includegraphics[width = 0.7\textwidth]{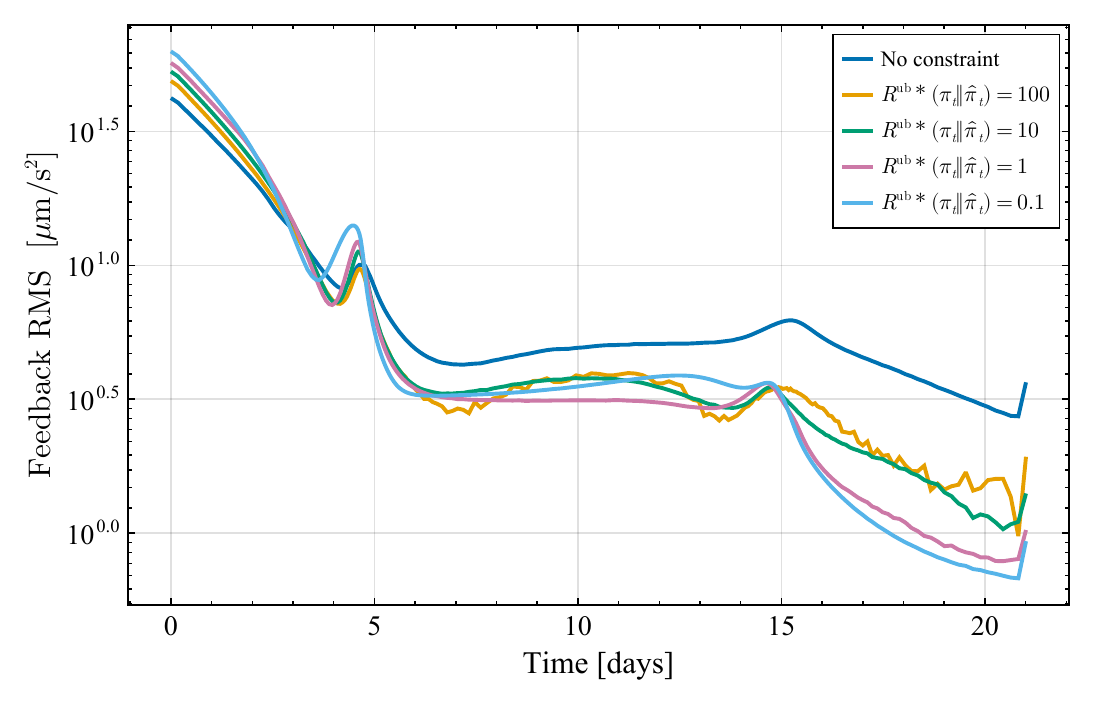}
    \caption{Feedback control RMS over the scenario duration. The solid blue line indicates the non-constraint case. Each subsequent line indicates a case with a relative entropy upper bound imposed.}
    \label{fig:tc1_feedback_RMS}
\end{figure}

We test a set of ${\mathcal{R}^{\text{ub}}_N}^*$, spaced in powers of ten from 100 to 0.1. In Figure \ref{fig:tc1_entropy} we show the solved-for relative entropy upper bound generated over the scenario duration. The dark blue line corresponds to no upper bound constraint. Each subsequent solid line corresponds to a case with a constraint whose value is indicated by the dashed line of the same color. There are common characteristics across each of the constraint values in the test set. Of particular note is the relatively rapid increase between days 3 and 5 of the scenario -- corresponding to periapse passage of the nominal transfer. The rapid increase in the bound coincides with the nonlinearities in this regime that accelerate divergence between the true distribution, $\pi_t$, and the Gaussian reference, $\hat{\pi}_t$. In Figure \ref{fig:tc1_feedback_RMS} the feedback RMS plotted over the duration of the scenario. During the initial leg, there is a large increase in the expected control variance for all cases. Subsequently, there are two smaller spikes -- one at about 4 days and one at 15 days -- corresponding to perilune passages. Generally, imposing a tighter relative entropy constraint yields a larger control variance. 

% TC1 covariance trajectory -- nominal
\begin{figure}[htb!]
    \centering
    \includegraphics[width = 0.60\textwidth]{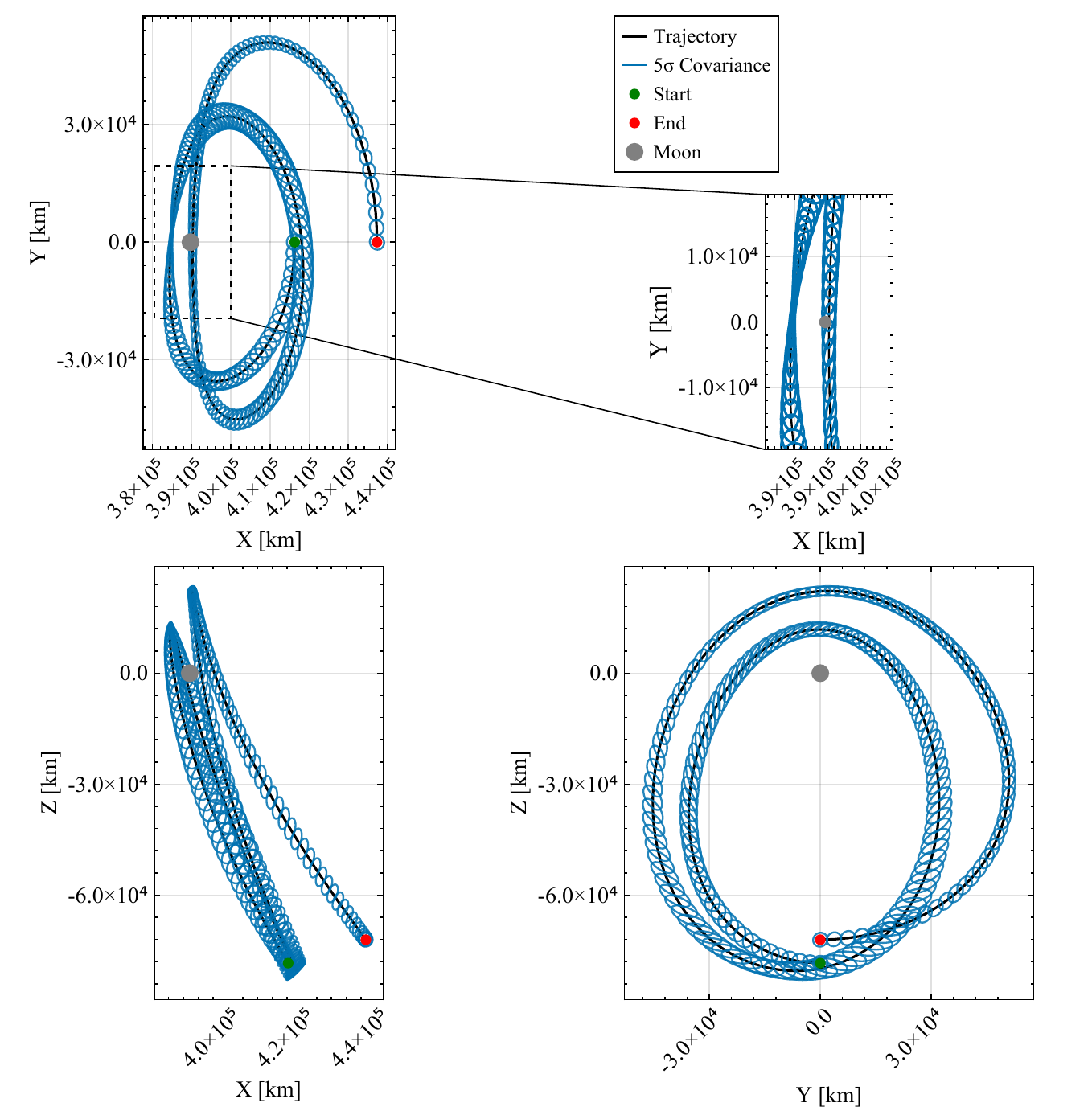}
    \caption{Covariance evolution using standard covariance steering. The blue ellipses indicate the solved-for $5\sigma$ covariance at each of the discrete nodes.}
    \label{fig:tc1_covariance_trajectory_nominal}
\end{figure}

% TC1 covariance trajectory -- constrained
\begin{figure}[htb!]
    \centering
    \includegraphics[width = 0.60\textwidth]{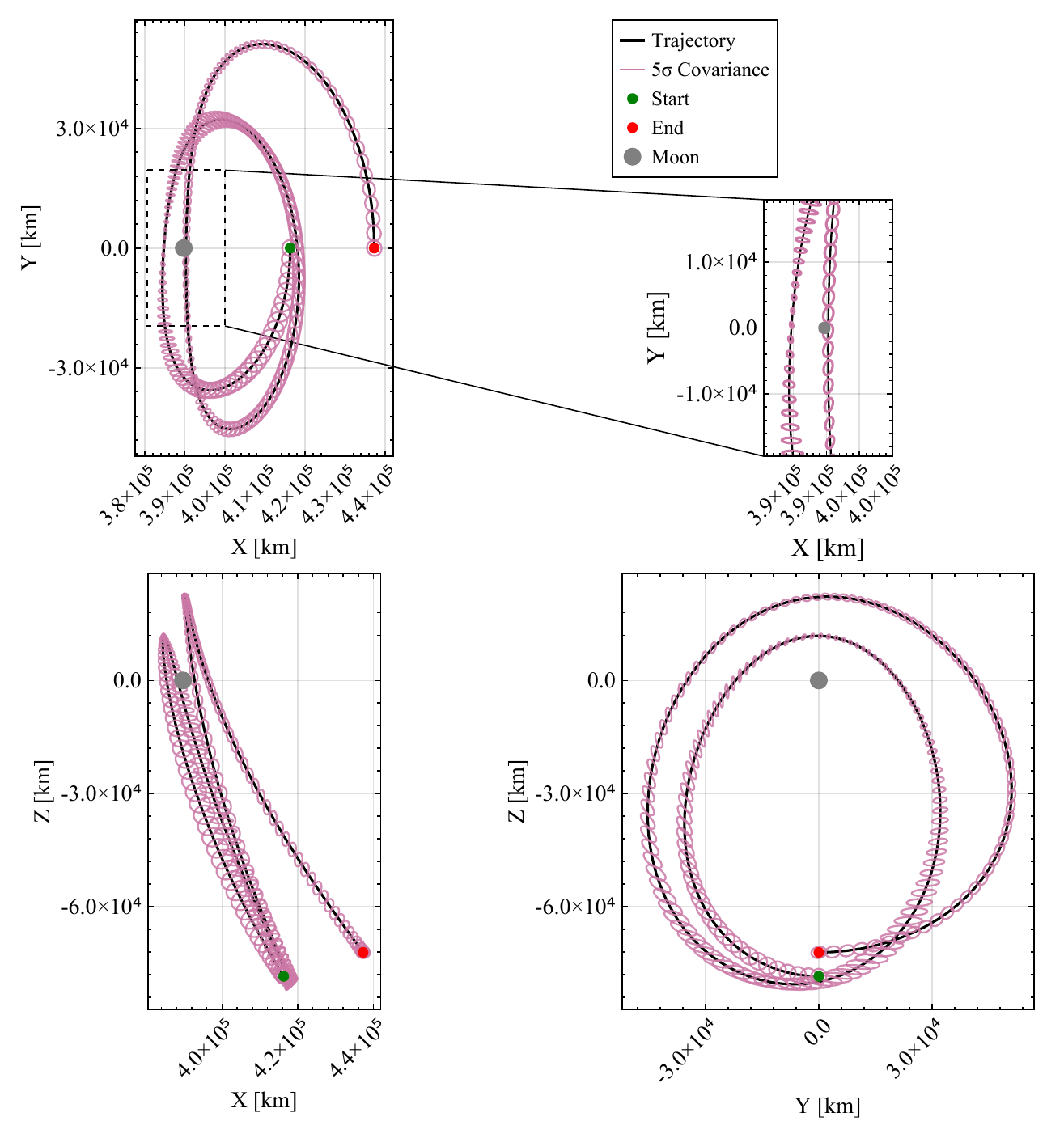}
    \caption{Covariance evolution using the relative entropy constrained covariance steering approach proposed. The purple ellipses indicates the solved-for $5\sigma$ covariance at each of the discrete nodes. }
    \label{fig:tc1_covariance_trajectory_constrained}
\end{figure}

In Figures \ref{fig:tc1_covariance_trajectory_nominal} and \ref{fig:tc1_covariance_trajectory_constrained} we show the solved-for covariance evolution for solutions generated with standard covariance steering, i.e., baseline, and the proposed relative entropy constrained method with ${\mathcal{R}_N^{\text{ub}}}^* = 1.0$, respectively. The panels in each set of subplots show the trajectory's and covariance evolution's projection onto each of the coordinate planes. In both sets, we highlight the portion in the $x-y$ plane closest to perilune. The standard covariance steering case, i.e., the blue ellipses, shows that close to perilune, the covariance ellipsoids are stretched in the along-track direction. However, when applying the relative entropy constraint, the guidance policy seeks to maintain orthogonality of the covariance principal axes with respect to these directions of maximum stretching, as is apparent in analyzing the purple ellipses. This result is also consistent with our analytical expectation: recall that
\begin{equation}
    \dot{\mathcal{R}}^{\text{ub}}_k(\pi_t\|\hat{\pi}_t) = \frac{1}{8\sigma^2} \sum_{i = 1}^n\left[\operatorname{tr}(H_k^i P_k)^2 + 2\operatorname{tr}\left( (H_k^i P_k)^2 \right) \right] \nonumber
\end{equation}
Therefore, to bound the growth in $\mathcal{R}^{\text{ub}}$, the guidance policy should utilize orthogonality of the decision variable $P_k$ with respect to $H_k^i$, prescribing the direction and magnitude of second-order nonlinearities. Additionally, it is noticeable that generally the entropy constrained solution exhibits tighter $5\sigma$ covariance bounds. By constraining the delivery error closer to the reference trajectory, the guidance policy limits linearization errors, and therefore damps the growth in relative entropy. By inspection, this is also consistent with the analytical form of the growth rate presented above. 

\begin{figure}[htb!]
    \centering
    \includegraphics[width = 0.8\textwidth]{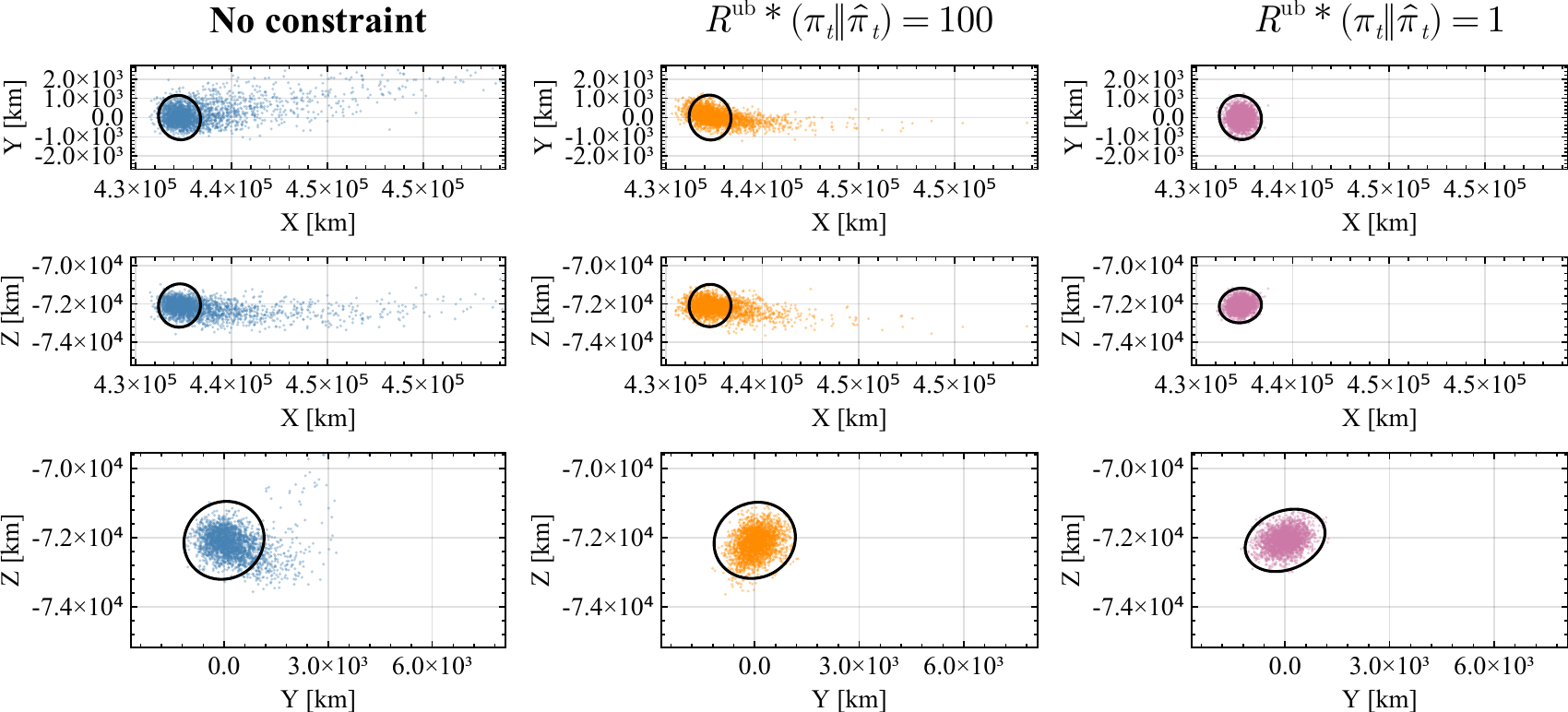}
    \caption{Each column shows a Monte Carlo analysis of 1) baseline covariance steering, 2) covariance steering constrained by $R_N^{\text{ub}} = 100.0$, and 3) covariance steering constrained by $R_N^{\text{ub}} = 1.0$. The black ellipses in each panel correspond to the solved-for $3\sigma$ covariance.}
    \label{fig:tc1_monte_carlo}
\end{figure}

We perform a Monte Carlo analysis by propagating the true state dynamics with stochastic accelerations and the solved-for guidance policy over $2500$ trials. To that end, the discrete-time Monte Carlo dynamics are 
\begin{equation}
    \boldsymbol{x}_{k + 1} = \boldsymbol{f}_k(\boldsymbol{x}_k, \boldsymbol{u}_k, \boldsymbol{u}_{k + 1}) + \boldsymbol{w}_k, \nonumber
\end{equation}
where 
\begin{equation}
    \boldsymbol{f}_k(\boldsymbol{x}_k, \boldsymbol{u}_k, \boldsymbol{u}_{k + 1}) = \int_{t_k}^{t_{k + 1}} \boldsymbol{f}(\boldsymbol{x}_\tau, \boldsymbol{u}_\tau,  \tau)d\tau \nonumber,
\end{equation}
and
\begin{equation}
    \boldsymbol{u}_t = \frac{t_{k + 1} - t}{t_{k + 1} - t_{k}}\boldsymbol{u}_k + \frac{t - t_k}{t_{k + 1} - t_k} \boldsymbol{u}_{k + 1} \nonumber.
\end{equation}
Recall that $\boldsymbol{u}_k = \boldsymbol{u}_k^* + K_k\tilde{\boldsymbol{x}}_k$, where $K_k$ is the solved-for guidance feedback gain matrix. The process noise vector $\boldsymbol{w}_k \sim \mathcal{N}(\boldsymbol{w}_k; 0, D_k)$. The results of this analysis are shown in Figure \ref{fig:tc1_monte_carlo}. The left-most column of panels shows the Monte Carlo points, evaluated at the terminal node for the standard covariance steering case. The black ellipses indicate the solved-for $3\sigma$ covariance. The center and right shown the same results, but for the cases where $\mathcal{R}_N^{\text{ub}} = 100.0$ and  $\mathcal{R}_N^{\text{ub}} = 1.0$, respectively. The left-most column shows a substantial number of points outside the $99\%$ confidence bound, indicating that standard covariance steering fails to produce consistent guidance for this test case. On the other hand, by imposing progressively tighter bounds for $\mathcal{R}_N^{\text{ub}}$, these results indicate that we can achieve stochastic guidance policies consistent with the designed-for terminal uncertainty bounds. 

% TC1 monte carlo

\subsection{Test Case \#2: \ac{MSE} Variational Upper Bound Constraints}

In this case, we impose the variational upper bound constraint for the \ac{MSE}, provided by Eq. \ref{eqn:MSE_upper_bound}. The bound value and other relevant parameters for the case are included in Table \ref{tab:conditions_case_12}. We solve the subproblem stated in Eq. \ref{eqn:subproblem_form_2} iteratively using the \ac{SCvx} algorithm to generate the guidance feedback policy. 

\begin{figure}[htb!]
    \centering
    \includegraphics[width = 1.0\textwidth]{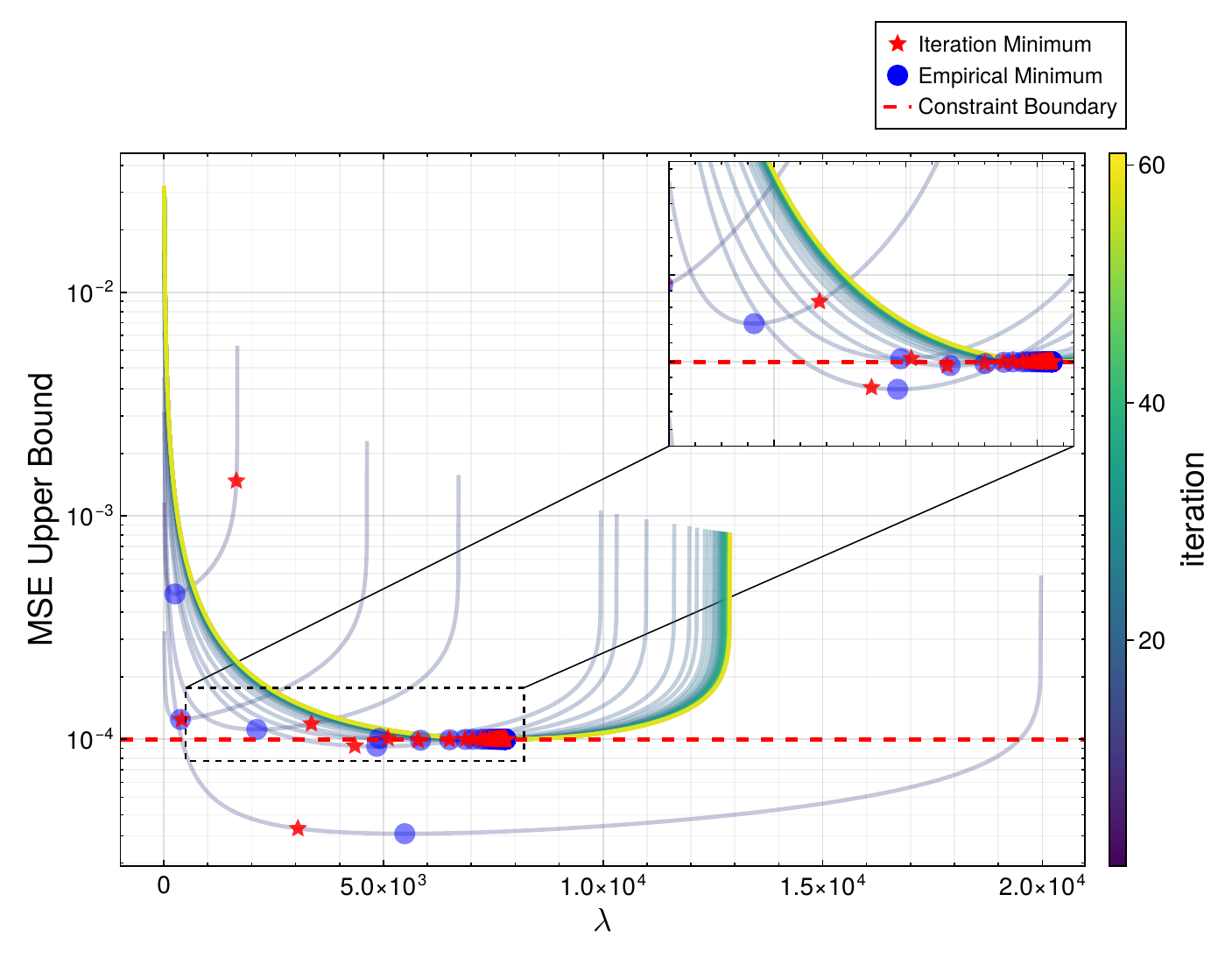}
    \caption{Convergence of $\lambda_N$ through successive convex iterations. Each curve shows Eq. \ref{eqn:MSE_upper_bound} evaluated at the iterate's solution for $P_N$ and $\mathcal{R}^{\text{ub}}_N$, generated over the range of feasible $\lambda_N$. The red points show the iterate's solved-for $\lambda_N$ and the blue indicate the true minimum.}
    \label{fig:tc2_gamma_hist}
\end{figure}

In Figure \ref{fig:tc2_gamma_hist} we show convergence of the first-order necessary condition for $\lambda^*$ using \ac{SCvx}. Each of the curves in the figure is Eq. \ref{eqn:MSE_upper_bound} evaluated at the iterate's solution for $P_N$ and $\mathcal{R}^{\text{ub}}_N$ over the range of feasible $\lambda_N$. The red and blue points on each curve indicate the iterate's solution for $\lambda_N$ and the true minimum of the curve, respectively. We can see that through successive iterations, the red and blue points converge, indicating that \ac{SCvx} successfully finds $\lambda^*_N$. Furthermore, the final bound evaluated at $\lambda^*$ straddles the \ac{MSE} bound constraint -- the red dashed line -- implying that $P_N$ and $\mathcal{R}_N^{\text{ub}}$ are optimal.  

\begin{figure}[htb!]
    \centering
    \includegraphics[width = 0.75\textwidth]{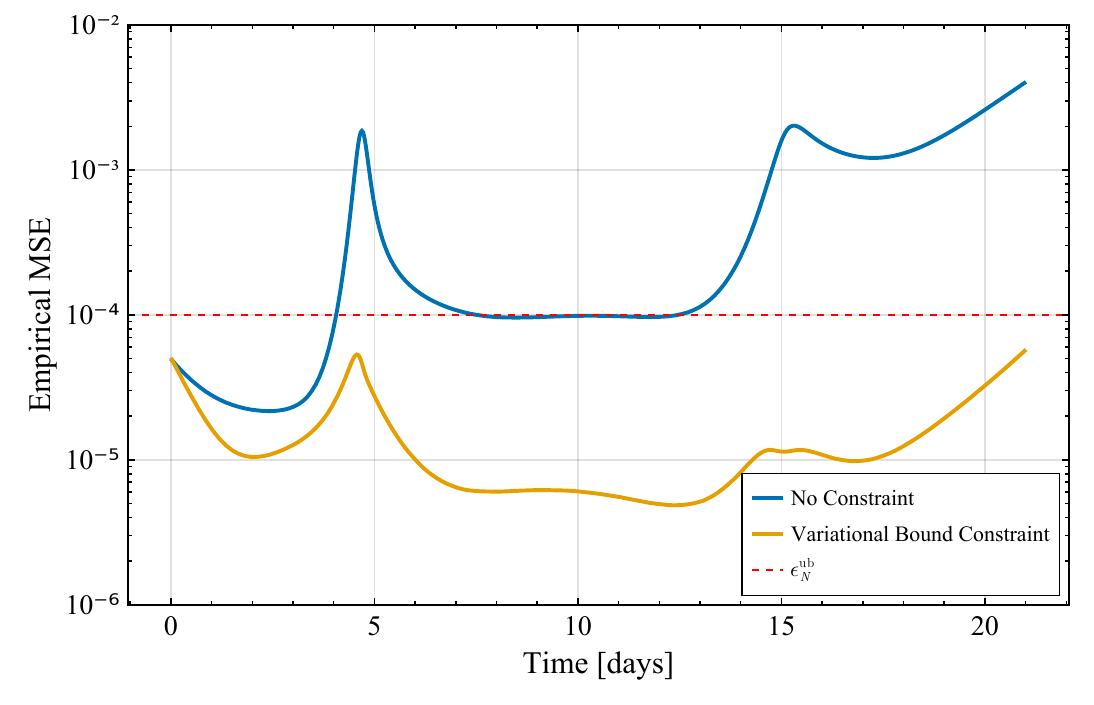}
    \caption{Empirical \ac{MSE} evaluated over the scenario duration for standard covariance steering and with the variational \ac{MSE} upper bound imposed. The red dashed line indicates the bound constraint.}
    \label{fig:tc2_mse_hist}
\end{figure}

We perform a Monte Carlo analysis to validate the efficacy of the variational upper bound proposed for constraining the true \ac{MSE}. Like the previous test case, 2500 trials are conducted using the procedure described above. We evaluate the empirical \ac{MSE} at each of the discrete nodes as 
\begin{equation}
    \widehat{\text{MSE}}_k = \frac{1}{N_{\text{trial}}} \sum_{i = 1}^{N_{\text{trial}}} \| \boldsymbol{x}_k^i - \boldsymbol{x}_k^*\|_2^2 \nonumber. 
\end{equation}
This result is plotted in Figure \ref{fig:tc2_mse_hist}. The standard covariance steering approach clearly does not obey the prescribed bound for the true \ac{MSE}. On the other hand, while somewhat conservative, the proposed guidance design successfully constrains the true \ac{MSE} to fall below the constraint value.

% --------------------------
% ------ Conclusions -------
% --------------------------

\section{Conclusions}
This work developed a distributionally robust stochastic guidance framework for nonlinear spacecraft systems using relative entropy to quantify the mismatch between a true state distribution and a Gaussian surrogate. By exploiting the Donsker-Varadhan variational formula, we derived upper bounds on risk-sensitive quantities, including collision risk and mean-squared error, over a KLD ambiguity set. We also established a bound on the time rate of change of the KLD and showed that, under a Gaussian reference approximation, this bound can be incorporated within a covariance-steering formulation. Embedding these results in a successive convexification algorithm enabled the design of feedback policies that simultaneously regulate distributional mismatch and enforce conservative bounds on risk-sensitive performance. Numerical results for a nonlinear transfer between near-rectilinear halo orbits demonstrate the practicality of the proposed approach and its ability to provide additional robustness beyond conventional Gaussian covariance steering. The framework offers a principled alternative to selecting ambiguity-set radii heuristically by connecting the radius directly to nonlinear dynamical evolution. Future work should investigate tighter KLD-rate bounds, alternative reference distributions, and incorporating model uncertainty. 

% --------------------------------------- %
% ------------ References --------------- %
% --------------------------------------- %

\clearpage
\bibliographystyle{AAS_publication} 
\bibliography{references}

\end{document}